\documentclass{article}

\usepackage[final, nonatbib]{nips_2018}

\usepackage{multirow}
\usepackage{graphicx}
\usepackage[utf8]{inputenc} 
\usepackage[T1]{fontenc}    
\usepackage{hyperref}       
\usepackage{url}            
\usepackage{booktabs}       
\usepackage{amsfonts}       
\usepackage{nicefrac}       
\usepackage{microtype}      
\usepackage[backend=biber,style=numeric,maxbibnames=9,maxcitenames=2]{biblatex}
\title{CLEAR: A Dataset for Compositional Language and Elementary Acoustic Reasoning}

\author{
  Jerome Abdelnour\\
  NECOTIS, ECE Dept.\\
  Sherbrooke University\\
  Québec, Canada \\
  \texttt{Jerome.Abdelnour}\\
  \texttt{@usherbrooke.ca}
\And
 Giampiero Salvi\\
 KTH Royal Institute\\ of
 Technology\\
  EECS School\\
  Stockholm, Sweden \\
  \texttt{giampi@kth.se}
  \And
  Jean Rouat\\
  NECOTIS, ECE Dept.\\
  Sherbrooke University\\
  Québec, Canada \\
  \texttt{Jean.Rouat}\\
  \texttt{@usherbrooke.ca} \\
}
\begin{document}
\maketitle

\vspace{-3mm}

\begin{abstract}
We introduce the task of acoustic question answering (AQA) in the area of acoustic reasoning. In this task an agent learns to answer questions on the basis of acoustic context. In order to promote research in this area, we propose a data generation paradigm adapted from CLEVR~\cite{johnson2017clevr}. We generate acoustic scenes by leveraging a bank of elementary sounds. We also provide a number of functional programs that can be used to compose questions and answers that exploit the relationships between the attributes of the elementary sounds in each scene. We provide AQA datasets of various sizes as well as the data generation code. As a preliminary experiment to validate our data, we report the accuracy of current state of the art visual question answering models when they are applied to the AQA task without modifications. Although there is a plethora of question answering tasks based on text, image or video data, to our knowledge, we are the first to propose answering questions directly on audio streams. We hope this contribution will facilitate the development of research in the area.
\end{abstract}

\section{Introduction and Related Work}

Question answering (QA) problems have attracted increasing interest in the machine learning and artificial intelligence communities.
These tasks usually involve interpreting and answering text based questions in the view of some contextual information, often expressed in a different modality.
Text-based QA, use text corpora as context (\cite{voorhees1999trec, voorhees2000building, soubbotin2001patterns, hovy2000question, iyyer2014neural,ravichandran2002learning}); in visual question answering (VQA), instead, the questions are related to a scene depicted in still images (e.g. \cite{johnson2017clevr,antol2015vqa,zhu2016visual7w,gao2015you, agrawal2016analyzing, zhang2016yin, geman2015visual, iyyer2014neural,ravichandran2002learning}. Finally, video question answering attempts to use both the visual and acoustic information in video material as context (e.g. \cite{cao2005automated,chua2003question,yang2003videoqa,kim2017deepstory,tapaswi2016movieqa,wu2008robust}). In the last case, however, the acoustic information is usually expressed in text form, either with manual transcriptions (e.g. subtitles) or by automatic speech recognition, and is limited to linguistic information~\cite{ZhangEtAl2017VQAwithspeech}.

The task presented in this paper differs from the above by answering questions directly on audio streams. We argue that the audio modality contains important information that has not been exploited in the question answering domain. This information may allow QA systems to answer relevant questions more accurately, or even to answer questions that are not approachable from the visual domain alone. Examples of potential applications are the detection of anomalies in machinery where the moving parts are hidden, the detection of threatening or hazardous events, industrial and social robotics.

Current question answering methods require large amounts of annotated data. In the visual domain, several strategies have been proposed to make this kind of data available to the community \cite{johnson2017clevr,antol2015vqa,zhu2016visual7w,gao2015you}. \citeauthor{agrawal2016analyzing} \cite{agrawal2016analyzing} noted that the way the questions are created has a huge impact on what information a neural network uses to answer them (this is a well known problem that can arise with all neural network based systems). This motivated research \cite{zhang2016yin, geman2015visual, johnson2017clevr} on how to reduce the bias in VQA datasets. The complexity around gathering good labeled data forced some authors \cite{zhang2016yin, geman2015visual} to constrain their work to yes/no questions. \citeauthor{johnson2017clevr} \cite{johnson2017clevr} made their way around this constraint by using synthetic data. To generate the questions, they first generate a semantic representation that describes the reasoning steps needed in order to answer the question. This gives them full control over the labelling process and a better understanding of the semantic meaning of the questions. They leverage this ability to reduce the bias in the synthesized data. For example, they ensure that none of the generated questions contains hints about the answer.

Inspired by the work on CLEVR~\cite{johnson2017clevr}, we propose an acoustical question answering (AQA) task by defining a synthetic dataset that comprises audio scenes composed by sequences of elementary sounds and questions relating properties of the sounds in each scene. We provide the adapted software for AQA data generation as well as a version of the dataset based on musical instrument sounds. We also report preliminary experiments using the FiLM architecture derived from the VQA domain.

\section{Dataset} \label{sec:dataset}
This section presents the dataset and the generation process\footnote{Available at \url{https://github.com/IGLU-CHISTERA/CLEAR-dataset-generation}}. In this first version (version 1.0) we created multiple instances of the dataset with 1000, 10000 and 50000 \emph{acoustic scenes} for which we generated 20 to 40 questions and answers per scene. In total, we generated six instances of the dataset. To represent questions, we use the same semantic representation through functional programs that is proposed in~\cite{johnson2017clevr,johnson2017inferring}. 

\newcommand{\emphPH}[1]{\textit{\textbf{#1}}}

\begin{table*}
    \centering
    \setlength{\tabcolsep}{4pt}
    \scriptsize
    \begin{tabular}{lp{0.52\textwidth}p{0.28\textwidth}c}
    \hline\hline
    Question type & Example & Possible Answers & \# \\
    \hline
    Yes/No       & Is there an equal number of \emphPH{loud} \emphPH{cello} sounds and \emphPH{quiet} \emphPH{clarinet} sounds? & yes, no & 2\\
    Note         & What is the note played by the \emphPH{flute} that is \emphPH{after} the \emphPH{loud} \emphPH{bright} \emphPH{D} note? & A, A\#, B, C, C\#, D, D\#, E, F, F\#, G, G\# & 12\\
    Instrument   & What instrument plays a \emphPH{dark} \emphPH{quiet} sound in the \emphPH{end} of the scene? & cello, clarinet, flute, trumpet, violin & 5 \\
    Brightness & What is the brightness of the \emphPH{first} \emphPH{clarinet} sound? & bright, dark & 2 \\
    Loudness & What is the loudness of the \emphPH{violin} playing after the \emphPH{third} \emphPH{trumpet}? & quiet, loud & 2 \\
    Counting     & How many other sounds have the same brightness as the \emphPH{third} \emphPH{violin}? & 0--10 & 11 \\
    Absolute Pos. & What is the position of the \emphPH{A\#} note playing after the \emphPH{bright} \emphPH{B} note? & \multirow{2}{*}{$\big\}$ first--tenth} & \multirow{2}{*}{10} \\
    Relative Pos. & Among the \emphPH{trumpet} sounds which one is a \emphPH{F}? &  &  \\
    Global Pos. & In what part of the scene is the \emphPH{clarinet} playing a \emphPH{G} note that is \emphPH{before} the \emphPH{third} \emphPH{violin} sound? & beginning, middle, end (of the scene) & 3 \\
    \hline
    Total & & & 47 \\
    \hline\hline
    \end{tabular}
    \caption{Types of questions with examples and possible answers. The variable parts of each question is emphasized in bold italics. In the dataset many variants of questions are included for each question type, depending on the kind of relations the question implies. The number of possible answers is also reported in the last column. Each possible answer is modelled by one output node in the neural network. Note that for absolute and relative positions, the same nodes are used with different meanings: in the first case we enumerate all sounds, in the second case, only the sounds played by a specific instrument.}
    \label{tab:types_of_question}
\end{table*}

\subsection{Scenes and Elementary Sounds} \label{sec:primary_sounds}
An acoustic scene is composed by a sequence of \emph{elementary sounds}, that we will call just sounds in the following. The sounds are real recordings of musical notes from the Good-Sounds database~\cite{goodSounds2016}. 
We use five families of musical instruments: cello, clarinet, flute, trumpet and violin.
Each recording of an instrument has a different musical note (pitch) on the MIDI scale. 
The data generation process, however, is independent of the specific sounds, so that future versions of the data may include speech, animal vocalizations and environmental sounds.
Each sound is described by an n-tuple \texttt{[Instrument family, Brightness, Loudness, Musical note, Absolute Position, Relative Position, Global Position, Duration]} (see Table~\ref{tab:types_of_question} for a summary of attributes and values).
Where \texttt{Brightness} can be either \texttt{bright} or \texttt{dark}; \texttt{Loudness} can be \texttt{quiet} or \texttt{loud}; \texttt{Musical note} can take any of the 12 values on the fourth octave of the Western chromatic scale\footnote{For this first version of CLEAR the cello only includes 8 notes: C, C\#, D, D\#, E, F, F\#, G.}. The \texttt{Absolute Position} gives the position of the sound within the acoustic scene (between first and tenth), the \texttt{Relative Position} gives the position of a sound relatively to the other sounds that are in the same category (e.g. ``the third cello sound''). \texttt{Global Position} refers to the approximate position of the sound within the scene and can be either \texttt{beginning}, \texttt{middle} or \texttt{end}.

\begin{figure}
    \centering
    \includegraphics[width=\textwidth]{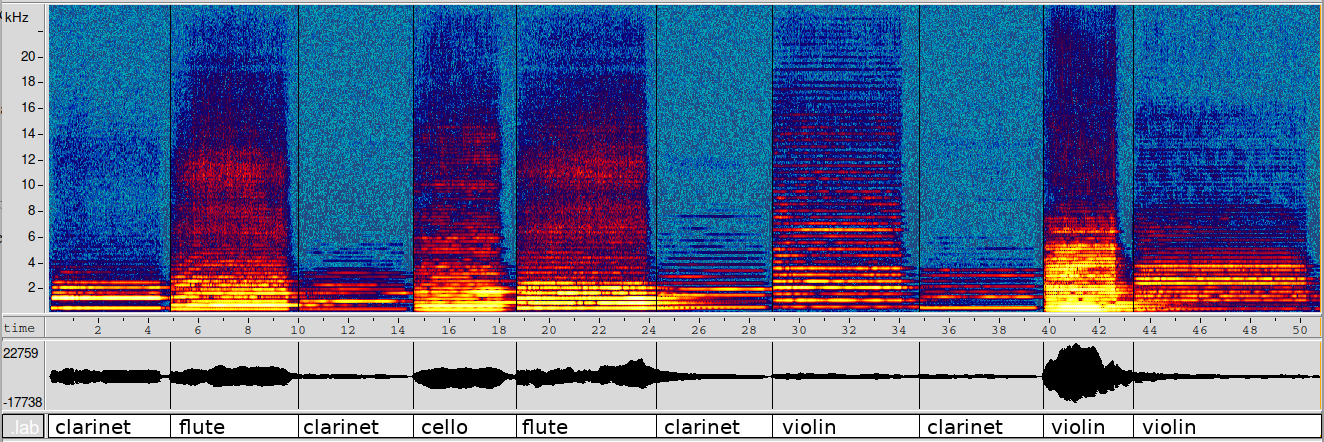}
    \caption{Example of an acoustic scene. We show the spectrogram, the waveform and the annotation of the instrument for each elementary sounds. A possible question on this scene could be "What is the position of the flute that plays after the second clarinet?", and the corresponding answer would be "Fifth". Note that the agent must answer based on the spectrogram (or waveform) alone.}
    \label{fig:spectrogram}
\end{figure}

We start by generating a \emph{clean} acoustic scene as following: first the encoding of the original sounds (sampled at 48kHz) is converted from 24 to 16 bits. Then silence is detected and removed when the energy, computed as $10 \log_{10}\sum_{i} x_{i}^{2}$ over windows of 100 msec, falls below -50 dB, where $x_i$ are the sound samples normalized between $\pm 1$. Then we measure the perceptual loudness of the sounds in dB~LUFS using the method described in the ITU-R BS.1770-4 international normalization standard~\cite{ITULoudness2015} and implemented in \cite{pyloudnorm}. We attenuate sounds that are in an intermediate range of -24~dB~LUFS and -30.5~dB~LUFS by -10~dB, to increase the separation between loud and quiet sounds. We obtain a bank of 56 elementary sounds. Each clean acoustic scene is generated by concatenating 10 sounds chosen randomly from this bank.

Once a clean acoustic scene has been created it is post-processed to generate a more difficult and realistic scene. A white uncorrelated uniform noise is first added to the scene. The amplitude range of the noise is first set to the maximum values allowed by the encoding. Then the amplitude is attenuated by a factor $f$ randomly sampled from a uniform distribution between -80~dB and -90~dB ($20 \log_{10} f$). The noise is then added to the scene. Although the noise is weak and almost imperceptible to the human ear, it guaranties that there is no pure silence between each elementary sounds. 
The scene obtained this way is finally filtered to simulate room reverberation using SoX\footnote{http://sox.sourceforge.net/sox.html}. For each scene, a different room reverberation time is chosen from a uniform distribution between [50ms, 400ms].

\subsection{Questions} \label{sec:questions}

Questions are structured in a logical tree introduced in CLEVR~\cite{johnson2017clevr} as a \emph{functional program}.A functional program, defines the reasoning steps required to answer a question given a scene definition. We adapted the original work of Johnson~\textit{et al.}~\cite{johnson2017clevr} to our acoustical context by updating the function catalog and the relationships between the objects of the scene. For example we added the \emph{before} and \emph{after} temporal relationships.

In natural language, there is more than one way to ask a question that has the same meaning. For example, the question “Is the cello as loud as the flute?” is equivalent to “Does the cello play as loud as the flute?”. Both of these questions correspond to the same functional program even though their text representation is different. Therefore the structures we use include, for each question, a functional representation, and possibly many text representations used to maximize language diversity and minimize the bias in the questions. We have defined 942 such structures.

A template can be instantiated using a large number of combinations of elements. Not all of them generate valid questions. For example "Is the flute louder than the flute?" is invalid because it does not provide enough information to compare the correct sounds regardless of the structure of the scene. Similarly, the question “What is the position of the violin playing after the trumpet?” would be ill-posed if there are several violins playing after the trumpet. The same question would be considered degenerate if there is only one violin sound in the scene, because it could be answered without taking into account the relation “after the trumpet”. A validation process~\cite{johnson2017clevr} is responsible for rejecting both ill-posed and degenerate questions during the generation phase.

Thanks to the functional representation we can use the reasoning steps of the questions to analyze the results. This would be difficult if we were only using the text representation without human annotations. If we consider the kind of answer, questions can be organized into 9 families as illustrated in Table~\ref{tab:types_of_question}. For example, the question “What is the third instrument playing?” would translate to the “Query Instrument” family as its function is to retrieve the instrument's name.

On the other hand we could classify the questions based on the relationships they required to be answered. For example, "What is the instrument after the trumpet sound that is playing the C note?" is still a “query\_instrument” question, but compared to the previous example, requires more complex reasoning. The appendix reports and analyzes statistics and properties of the database.

\section{Preliminary Experiments} \label{experimentationsSection}
To evaluate our dataset, we performed preliminary experiments with a FiLM network~\cite{perez2017film}. It is a good candidate as it has been shown to work well on the CLEVR VQA task~\cite{johnson2017clevr} that shares the same structure of questions as our CLEAR dataset. To represent acoustic scenes in a format compatible with FiLM, we computed spectrograms (log amplitude of the spectrum at regular intervals in time) and treated them as images. Each scene corresponds to a fixed resolution image because we have designed the dataset to include acoustic scenes of the same length in time. The best results were obtained with a training on 35000 scenes and 1400000 questions/answers. It yields a 89.97\% accuracy on the test set that comprises 7500 scenes and 300000 questions. For the same test set a classifier choosing always the majority class would obtain as little as 7.6\% accuracy.

\section{Conclusion}
We introduce the new task of acoustic question answering (AQA) as a means to stimulate AI and reasoning research on acoustic scenes. We also propose a paradigm for data generation that is an extension of the CLEVR paradigm: The acoustic scenes are generated by combining a number of elementary sounds, and the corresponding questions and answers are generated based on the properties of those sounds and their mutual relationships. We generated a preliminary dataset comprising 50k acoustic scenes composed of 10 musical instrument sounds, and 2M corresponding questions and answers. We also tested the FiLM model on the preliminary dataset obtaining at best 89.97\% accuracy predicting the right answer from the question and the scene. Although these preliminary results are very encouraging, we consider this as a first step in creating datasets that will promote research in acoustic reasoning. The following is a list of limitations that we intend to address in future versions of the dataset.

\subsection{Limitations and Future Directions}
In order to be able to use models that were designed for VQA, we created acoustic scenes that have the same length in time. This allows us to represent the scenes as images (spectrograms) of fixed resolution. In order to promote models that can handle sounds more naturally, we should release this assumption and create scenes of variable lenghts.
Another simplifying assumption (somewhat related to the first) is that every scene includes an equal number of elementary sounds. This assumption should also be released in future versions of the dataset.
In the current implementation, consecutive sounds follow each other without overlap. In order to implement something similar to occlusions in visual domain, we should let the sounds overlap.
The number of instruments is limited to five and all produce sustained notes, although with different sound sources (bow, for cello and violin, reed vibration for the clarinet, fipple for the flute and lips for the trumpet). We should increase the number of instruments and consider percussive and decaying sounds as in drums and piano, or guitar. We also intend to consider other types of sounds (ambient and speech, for example) to increase the generality of the data. 
Finally, the complexity of the task can always be increased by adding more attributes to the elementary sounds, adding complexity to the questions, or introducing different levels of noise and distortions in the acoustic data.

\section{Acknowledgements}
We would like to acknowledge the NVIDIA Corporation for donating a number of GPUs, the Google Cloud Platform research credits program for computational resources. Part of this research was financed by the CHIST-ERA IGLU project, the CRSNG and Michael-Smith scholarships, and by the University of Sherbrooke.

\printbibliography

\clearpage
\appendix
\section{Statistics on the Data Set}
\begin{figure}
    \centering
    \includegraphics[width=\textwidth]{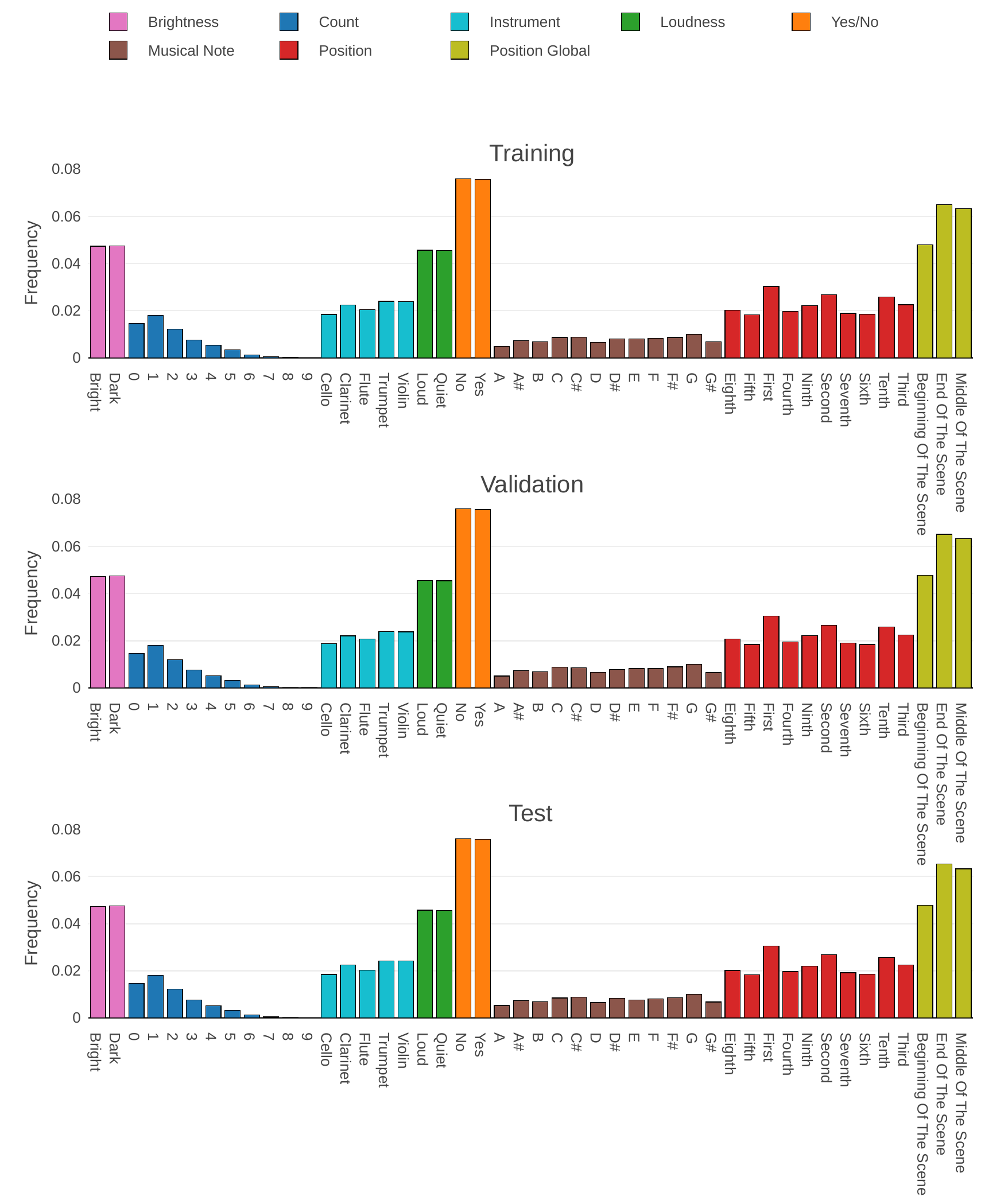}
    \caption{Distribution of answers in the dataset by set type. The color represent the answer category.}
    \label{fig:answer_distribution}
\end{figure}

\begin{figure}
    \centering
    \includegraphics[width=\textwidth]{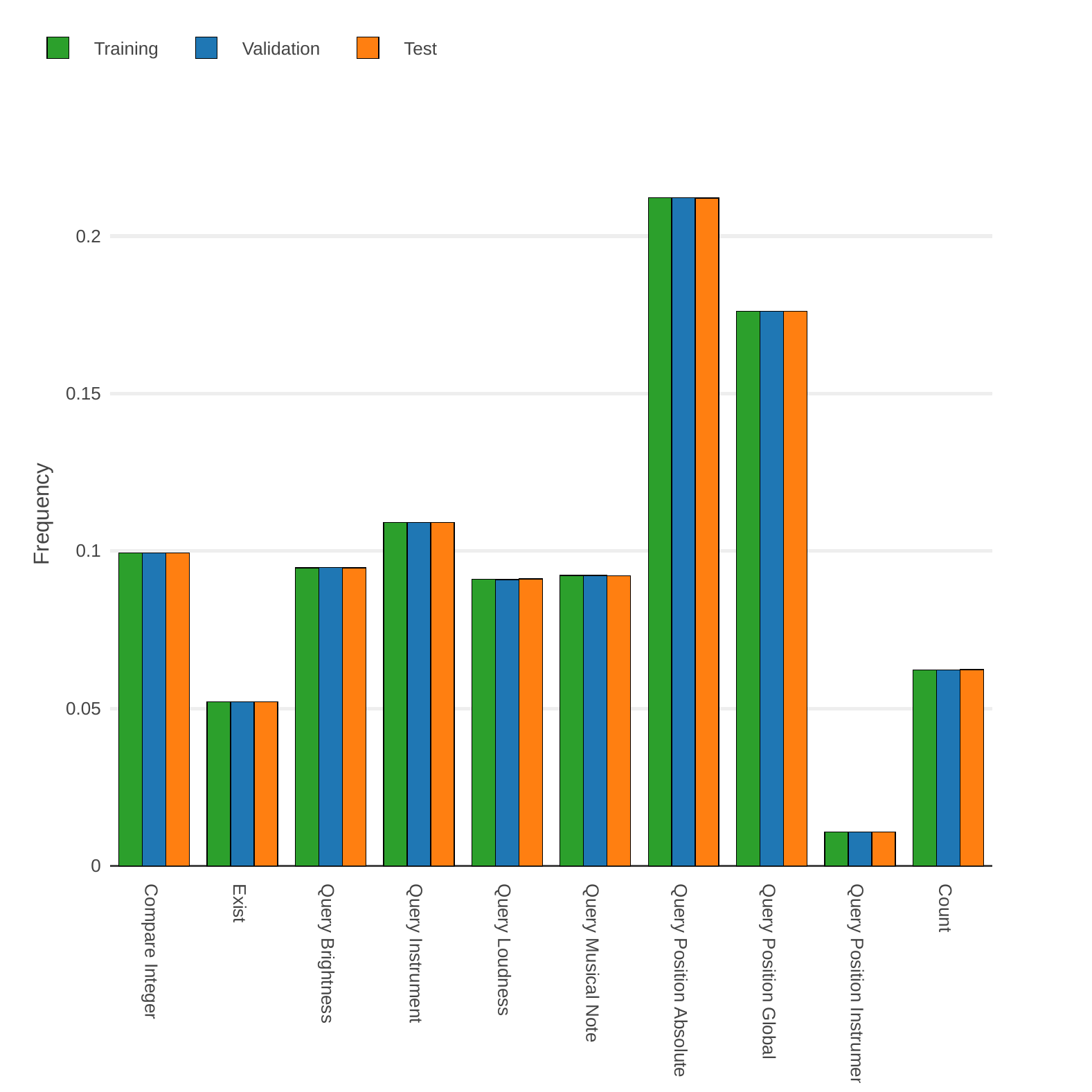}
    \caption{Distribution of question types. The color represent the set type.}
    \label{fig:question_type_distribution}
\end{figure}

\begin{figure}
    \centering
    \includegraphics[width=\textwidth]{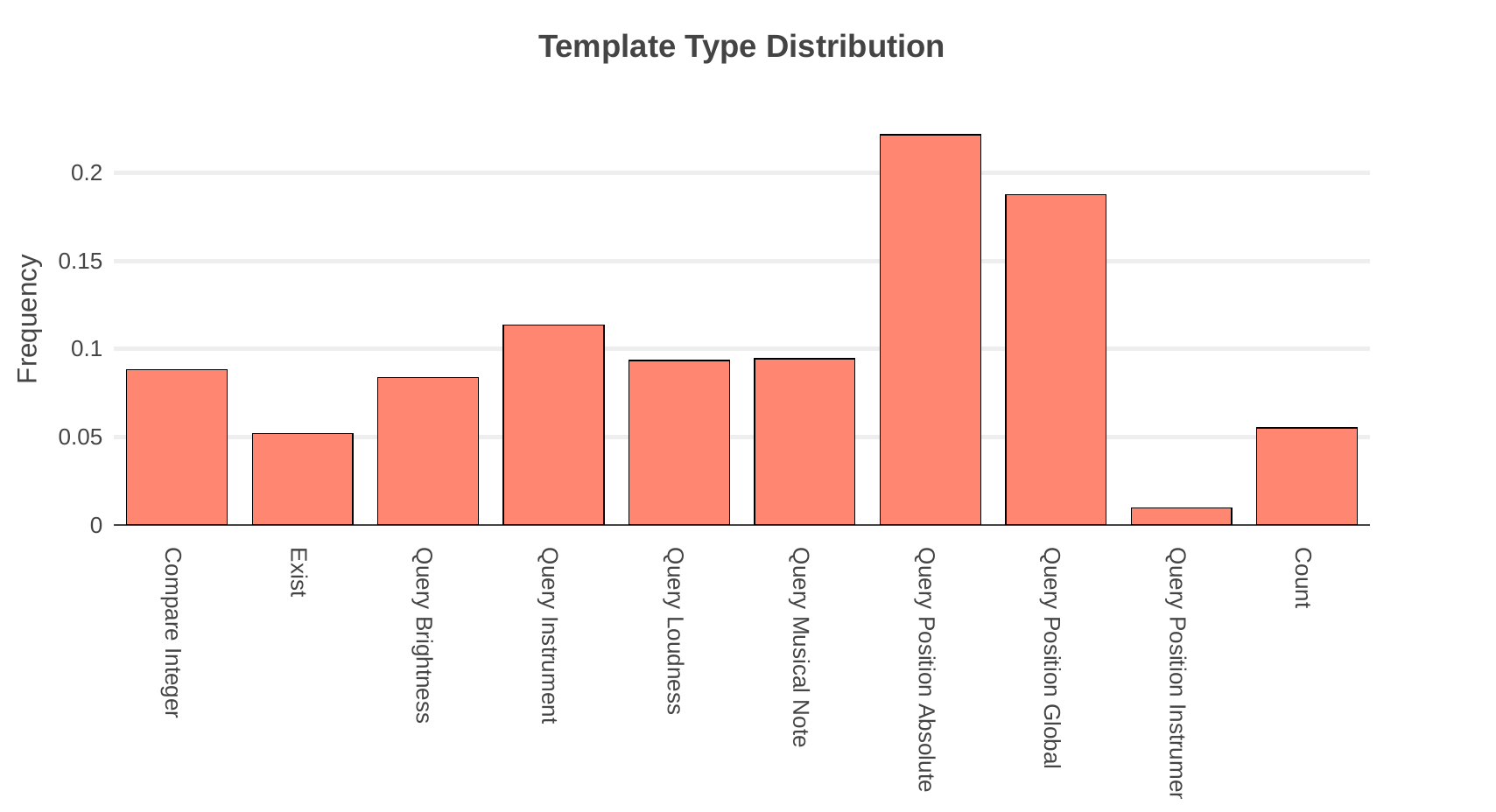}
    \caption{Distribution of template types. The same templates are used to generate the questions and answers for the training, validation and test set.}
    \label{fig:template_type_distribution}
\end{figure}

This appendix reports some statistics on the properties of the data set. We have considered the data set comprising 50k scenes and 2M questions and answers to produce the analysis. Figure~\ref{fig:answer_distribution} reports the distribution of the correct answer to each of the 2M questions. Figure~\ref{fig:question_type_distribution} and \ref{fig:template_type_distribution} reports the distribution of question types and available template types respectively. The fact that those two distributions are very similar means that the available templates are sampled uniformly when generating the questions.

\begin{figure}
    \centering
    \includegraphics[width=\textwidth]{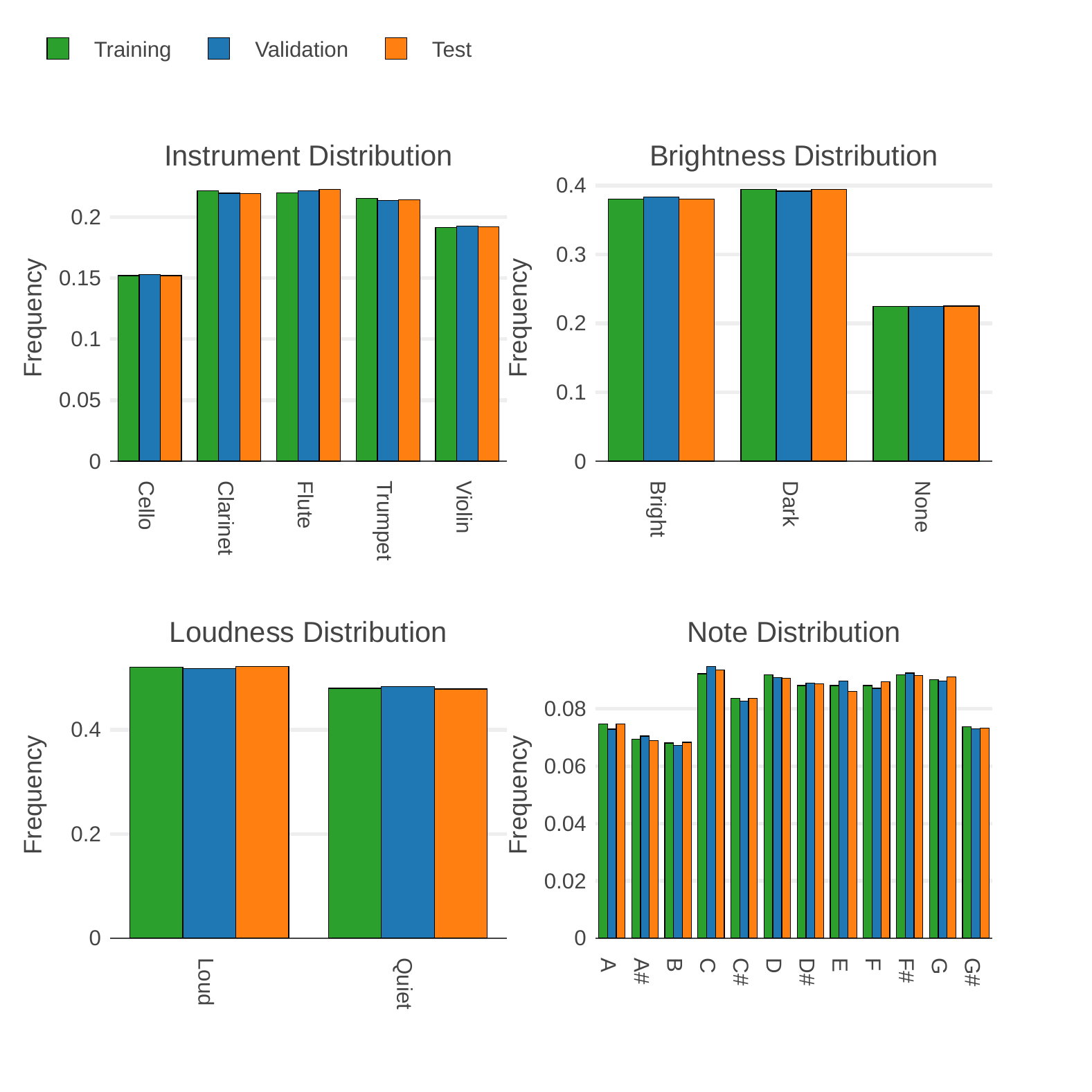}
    \caption{Distribution of sound attributes in the scenes. The color represent the set type. Sounds with a "None" brightness have an ambiguous brightness which couldn't be classified as 'Bright' or 'Dark'. }
    \label{fig:scene_distribution}
\end{figure}

Finally, Figure~\ref{fig:scene_distribution} shows the distribution of sound attributes in the scenes. It can be seen that most attributes are nearly evenly distributed. In the case of brightness, calculated in terms of spectral centroids, sounds were divided into clearly bright, clearly dark and ambiguous cases (referred to by "None" in the figure). We only instantiated questions about the brightness on the clearly separable cases.

\end{document}